\documentclass[letterpaper]{article} 
\usepackage{aaai25}  
\usepackage{times}  
\usepackage{helvet}  
\usepackage{courier}  
\usepackage[hyphens]{url}  
\usepackage{graphicx} 
\usepackage{natbib}  
\usepackage{caption} 
\usepackage{algorithm}
\usepackage{algorithmic}
\usepackage{amsmath}
\usepackage{amsfonts}
\usepackage{multirow}
\usepackage{svg}
\usepackage{booktabs}
\usepackage{pdfpages}

\usepackage{newfloat}
\usepackage{listings}
\DeclareCaptionStyle{ruled}{labelfont=normalfont,labelsep=colon,strut=off} 
\floatstyle{ruled}
\newfloat{listing}{tb}{lst}{}
\floatname{listing}{Listing}
\DeclareMathOperator*{\argmin}{arg\,min}

\title{ASER: Activation Smoothing and Error Reconstruction for Large Language Model Quantization}
\author{
    Weibo Zhao\equalcontrib, Yubin Shi\equalcontrib\thanks{Corresponding Author.}, Xinyu Lyu, Wanchen Sui, Shen Li, Yong Li 
}
\affiliations{
    Alibaba Cloud Computing \\
    \{weibo.zwb, shiyubin.syb\}@alibaba-inc.com
}

\begin{document}

\maketitle

\begin{abstract}
    Quantization stands as a pivotal technique for deploying large language models~(LLMs), yet it poses significant challenges particularly in achieving effective low-bit quantization. The limited numerical mapping makes the quantized model produce a non-trivial error, bringing out intolerable performance degradation. This paper is anchored in the basic idea of model compression objectives, and delves into the layer-wise error distribution of LLMs during post-training quantization. Subsequently, we introduce ASER, an algorithm consisting of (1)~\textbf{Error Reconstruction}: low-rank compensation for quantization error with LoRA-style matrices constructed by whitening SVD; (2)~\textbf{Activation Smoothing}: outlier extraction to gain smooth activation and better error compensation. ASER is capable of quantizing typical LLMs to low-bit ones, particularly preserving accuracy even in W4A8 per-channel setup. Experimental results show that ASER is competitive among the state-of-the-art quantization algorithms, showing potential to activation quantization, with minor overhead.
\end{abstract}

\section{Introduction}

The proliferation of large language models~(LLMs) places increasing demands on computation and storage~\cite{vaswani2017attention, devlin2018bert}, and stretches the boundaries of contemporary hardware prowess. Regarding typical generative models such as Llama3.1-310B~\cite{dubey2024llama3herdmodels} or Qwen2-72B~\cite{yang2024qwen2technicalreport}, they require hundreds of gigabytes of VRAM, and often rely on multi-GPU cluster data centers for deployment. Researchers have been trying to compress these large models while maintaining their performance in a more compact form. Quantization~\cite{frantar2022gptq, xiao2023smoothquant} of large language models emerges as a pivotal strategy in optimizing the deployment and computational efficiency of these heavy neural architectures. 

Quantization involves converting the model's weights and activations from floating-point numbers to lower-precision integers, whose primary goal is alleviating the computational and memory requirements without significantly compromising the model's performance, enabling edge deployment. Take post training quantization~(PTQ) as an example, current methods try to smooth~\cite{xiao2023smoothquant}, rotate~\cite{liu2024spinquant} or transform representations in models, showing promise of achieving quantization-friendly data range and remarkable compression rate. However, when it comes to low-bit quantization, even with the application of the aforementioned techniques, the error introduced by the quantized model becomes evident, resulting in notable degradation.

For typical quantization, the optimization objective can be formulated as minimizing $\|\textbf{WX}-\textbf{W}_q\textbf{X}\|_F$, where $\textbf{W}$ and $\textbf{W}_q=Q(\textbf{W})$ are original model weight and its quantized one respectively. We empirically find this inevitable quantization error has low-rank property, whose singular value distribution featuring a small number of high values and a long-tail bulk of low ones. This varies both intra and inter Transformer blocks, which prompts us to consider using a module similar to LoRA~\cite{hu2021lora} to reconstruct this error. Furthermore, we find the channels produce the major error are consistent with the existence of the outliers.

\begin{figure*}[t]
\centering
\includegraphics[width=1.88\columnwidth]{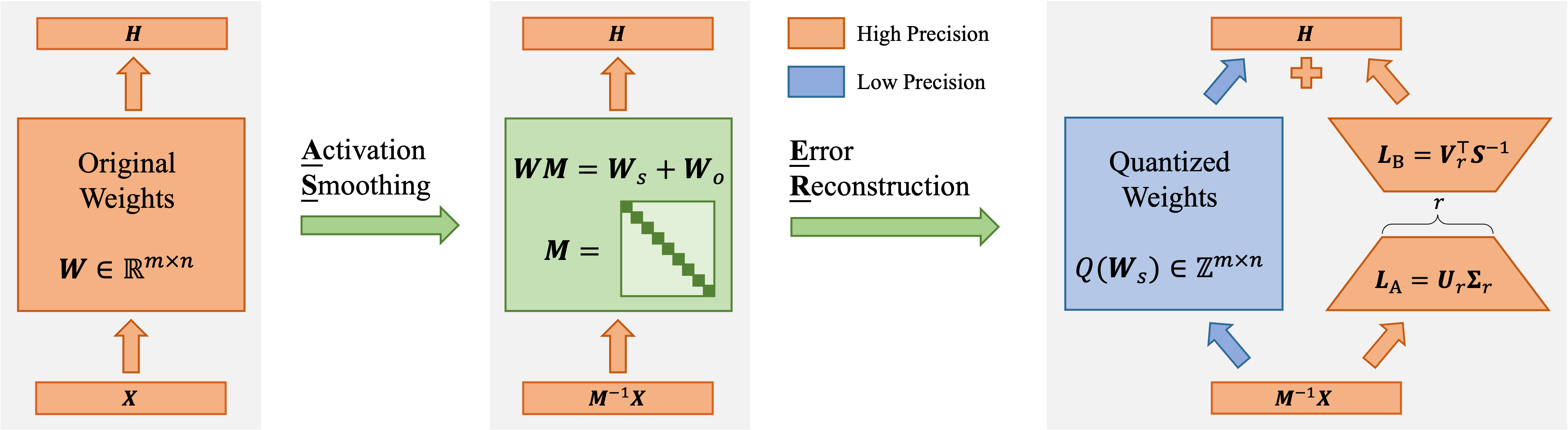}
\caption{The framework of ASER. The LoRA-style matrices $\textbf{L}_A, \textbf{L}_B$ are generated to reconstruct the quantization error.}
\label{fig:arch}
\end{figure*}

We propose ASER~(Activation Smoothing and Error Reconstruction), a low-rank compensation algorithm designed to enhance the efficacy of quantized models. Our method assesses the low-rank properties by whitening the relationship between quantization loss and its singular values based on Cholesky factorization. Then, ASER identifies these error-inducing outliers for smoothing. Finally, LoRA-style matrices are used to reconstruct the quantization error, with little computation overhead. The overall framework is shown in Figure~\ref{fig:arch}. This lightweight yet powerful mechanism ensures robust performance of quantized models, which is orthogonal to any particular weight quantization method. Our experiments demonstrate ASER remarkably recovers the quantization error, where W4A8 per-channel quantized model performs nearly equal capability compared to the half-precision reference. This work makes the following contributions:
\begin{itemize}
    \item We formulate the optimization objective of minimizing the discrepancy in model outputs, and analyze the characteristics of quantization errors in LLMs quantization.
    \item We propose a novel algorithm ASER, which includes error reconstruction using whitening SVD, and activation smoothing with outlier analysis.
    \item Experimental results show ASER can significantly recover the performance of quantized model in W4A8 per-channel quantization, with little computational overhead.
\end{itemize}

\section{Related Work}

\noindent\textbf{LLMs Compression and Quantization.} In recent years, given the powerful emergent capabilities~\cite{schaeffer2024emergent} of large language models~(LLMs), there has been a growing number of model compression techniques specifically developed for Transformer-based models. Network pruning~\cite{ma2023llm}, including structured~\cite{ma2023llm} and unstructured pruning~\cite{kurtic2022optimal}, removes unnecessary neurons or structures to decrease model size, while distillation~\cite{sun2020contrastive} directly uses a smaller model to learn the knowledge from the origin model. Though these conventional compression techniques have indeed yielded some benefits in LLMs, they pose challenges in deployment due to the modifications of model structure, and they often necessitate substantial consumption for retraining the models.

Post-Training Quantization~(PTQ) has emerged as a foundational technology for deploying large models, overcoming these limitations and significantly enhancing inference speed. GPTQ~\cite{frantar2022gptq}, as an extension of OBQ~\cite{frantar2022optimal}, employs second-order methods to obtain a closed-form solution, theoretically ensuring the minimization of quantization error. AWQ~\cite{lin2024awq} employs a scaling factor to reduce the quantization error of salient weights. But weight-only quantization methods do not consider decreasing the bit of activations to achieve further speedup. SmoothQuant~\cite{xiao2023smoothquant} uses empirical transformations to shift the quantization difficulty from activations to weights. However, it doesn't establish the relationship between transition matrix and quantization error, and pollutes the smoothness of weights.

\noindent\textbf{Low Rank Adapters of LLMs.} The low-rank property is a prominent characteristic of LLMs, and researchers have attempted to structurally compress models based on this observation. Low-rank decomposition is a critical technique for compression, which has been widely utilized in LLMs~\cite{wang2024svd}. Linformer~\cite{wang2020linformer} introduces a low-rank attention mechanism, reducing the original $O(N^2)$ computational complexity. For decoder models, DeepSeek-V2~\cite{deepseekai2024deepseekv2strongeconomicalefficient} exploits the low-rank property of KV cache to adopt the MLA architecture, which effectively compresses the storage of the KV cache. 

Another influential technique is LoRA~\cite{hu2021lora}, which fine-tunes large models, establishes a fundamental paradigm for large model fine-tuning. This approach of using low-rank matrices has been adopted across various domains. Recently, works like LoftQ~\cite{li2023loftq} and LQ-LoRA~\cite{guo2023lq} employ low-rank approximation to facilitate more efficient fine-tuning of lightweight models. Our focus lies in post-training quantization, aiming to maintain good performance when deploying low-bit models.
LoRC~\cite{yao2024exploring} employs them to improve the quality recovery of quantized models. L$^2$QER~\cite{zhang2024lqer} further scales the quantization error by empirically designed diagonal matrix before reconstruction.  We are inspired by these LoRA-style method and propose a more compact quantization framework from the perspective of model compression. ASER designs whitening SVD for error reconstruction and smoothes activations via outlier analysis.

\section{Analysis of Quantization Error}

In this section, we formulate the quantization error from a model compression perspective, and anchor minimizing the quantization loss as the objective of our problem. We first analyze the properties of the quantization loss in the context of RTN~(Round-To-Nearest) quantization, and then discuss how to reconstruct it with a small number of parameters.

\subsection{Minimizing the Quantization Error}

Tensors in quantized models have restricted range of representations, which brings out an unavoidable loss of dequantized data. In contrast to methods like LoRC that only account for the error arising from weight quantization, we aim to consider the integral error introduced by both the weights and the activations. Increasingly, quantization-related research~\cite{xiao2023smoothquant} acknowledges the impact of activations on quantization, primarily due to the critical role of outliers in the activations within LLMs~\cite{tang2024easyquantefficientdatafreequantization}. Thus, as a typical model compression, we formulate quantization objective optimization as:

\begin{equation}
    \argmin_{\textbf{W}_q}\|\textbf{WX}-\textbf{W}_q\textbf{X}\|_F,
\end{equation}

where $\textbf{W}_q=Q(\textbf{W})$ is the quantized weights, and we use Frobenius norm to measure the magnitude of the error matrix. Here, we use $\textbf{E}_q=\textbf{W}-\textbf{W}_q$ to represent the quantization error of the weight matrix. We aim to estimate it with an approximation $\tilde{\textbf{E}}_q$, and maximize the compensation for the overall quantization error $\tilde{\textbf{W}} = \textbf{W}_q + \tilde{\textbf{E}}_q$. Thus, the optimization objective can be transformed into:

\begin{equation}
   \argmin_{\tilde{\textbf{E}}_q}\|(\textbf{E}_q-\tilde{\textbf{E}}_q)\textbf{X}\|_F.
\end{equation}

\subsection{Characteristics of Quantization Error}

We conduct empirical studies on $\textbf{E}_q$ to guide our reconstruction strategy for the quantization error. We take Round-To-Nearest~(RTN) quantization on LLaMA3-8B~\cite{touvron2023llama} as an example to obtain the layer-wise distribution. 

\begin{figure}[h]
\centering
\includesvg[width=0.9\columnwidth]{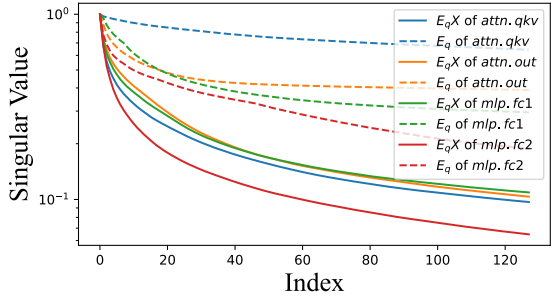}
\caption{The largest $128$ singular value distribution of quantization error in the $30^\text{th}$ layer of LLaMA3-8B by RTN quantization. The singular values are normalized for comparison.}
\label{fig:emp1}
\end{figure}

\noindent\textbf{Activation-weight quantization error exhibits low-rank properties.} As Figure~\ref{fig:emp1} shows, we plot the normalized singular value distribution of $\textbf{E}_q$ and $\textbf{E}_q\textbf{X}$ over four consecutive linear layers, i.e., $qkv\_proj$, $out\_proj$ of the multi-head self-attention~(MHSA), and $fc1$, $fc2$ of the feed-forward network~(FFN), in a certain Transformer block. We sort them and display the top $128$ largest ones out of $4096$ singular values. Compared to the weight error $\textbf{E}_q$, the distribution of activation-weight quantization error $\textbf{E}_q\textbf{X}$ features low-rank property, with a few larger values and a long tail with many smaller ones. This phenomenon is also consistent with the observation of~\citep{yu2023compressing} that the features in Transformers are low-rank rather than the weights. Therefore, we can leverage this property to estimate the quantization error with low-rank approximation, as the approach of low-rank adapters~\cite{hu2021lora}. Especially, in consideration of $\textbf{E}_q\textbf{X}$, we reconstruct the data-aware integral error rather than the weights themselves like~\citep{yao2024exploring} does. 

\begin{figure}[h]
\centering
\includesvg[width=0.9\columnwidth]{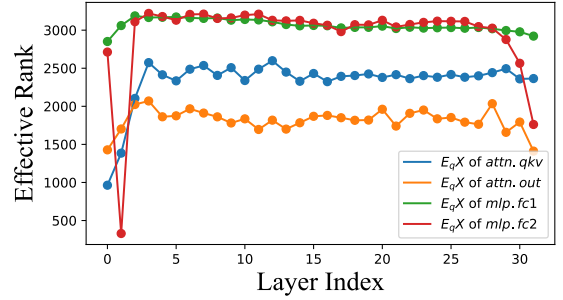}
\caption{The effective rank of $\textbf{E}_q\textbf{X}$ across layers in LLaMA3-8B by RTN quantization.}
\label{fig:emp2}
\end{figure}

\noindent\textbf{Low-rankness varies intra- and inter- layers.} We introduce a new metric to quantify the amount of effective dimensionality in a low-rank matrices, effective rank~\cite{roy2007effective}. Suppose the matrix $\textbf{Z}$ has singular values $\sigma_1, \sigma_2, ..., \sigma_n$, its effective rank can be computed as the entropy of the normalized singular values:

\begin{equation}
    \operatorname{Eff\_Rank}(\textbf{Z})=\exp (-\sum_{k=1}^{n} p_k \log p_k),
\label{equ:eff_rank}
\end{equation}
\begin{equation}
    p_k=\frac{\sigma_k(\textbf{Z})}{\sum_{i=1}^n \sigma_i(\textbf{Z})}+\epsilon.
\end{equation}

We plot the activation-weight quantization error $\textbf{E}_q\textbf{X}$ across layers in Figure~\ref{fig:emp2} according to the definition of effective rank in Equ.\ref{equ:eff_rank}. As we can see, the quantization error of linear layers in the MHSA tend to be more low-rank, whereas the counterparts in FFN have relatively higher effective dimensionality. Additionally, when examining the behavior across Transformer layers in LLaMA3-8B, we find that the quantization errors are more strongly low-rank in the layers closer to the input. This might be related to the smoother numerical distribution in the early layers of the model, which could contribute to this characteristic. Therefore, it is essential to adaptively choose appropriate ranks and the dimensions of narrow matrices to compensate for the quantization loss in each individual linear layer.

\begin{figure}[h]
\centering
\includesvg[width=0.9\columnwidth]{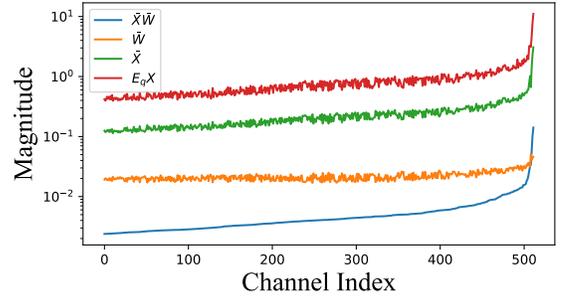}
\caption{Magnitude of activation-weight quantization error $\textbf{E}_q\textbf{X}$, mean activation $\bar {\textbf{X}}$, mean weight $\bar {\textbf{W}}$ and their product $\bar {\textbf{X}}\bar {\textbf{W}}$. The channel index is sorted by $\bar {\textbf{X}}\bar {\textbf{W}}$. Data comes from the largest $512$ channels of a linear layer in LLaMA3-8B when conduting RTN quantization.}
\label{fig:emp3}
\end{figure}

\noindent\textbf{Outliers constitute the major component of quantization error.} Increasing research~\cite{dettmers2022llmint88bitmatrixmultiplication, xiao2023smoothquant, ma2023llm} has recognized the significant impact of outliers on model quantization, making their handling a key consideration in quantization algorithms. In our experiments, we also observed that both activation and weight channel outliers contribute significantly to the large components in quantization error. As shown in Figure~\ref{fig:emp3}, when we sort the channels by $\bar {\textbf{X}}\bar {\textbf{W}}$, the Frobenius norm of the quantization error has also demonstrated a broadly similar trend. Notably, we find that a small fraction (less than 1\% of the 4096) of outlier channels are responsible for an order of magnitude more quantization error than the rest, and that these channels have large magnitude in both weights and activations. As a result, we attempt to implement targeted quantization and compensation strategies for them.

\section{ASER: Activation Smoothing and Error Reconstruction}

In this section, we introduce ASER, which consists of two main technologies. Error Reconstruction~(ER) constructs the compensation matrices using a whitening method whose singular values are directly related to the integral error. Activation Smoothing~(AS) analyzes outliers and transforms fluctuations in activation to weights, making low-rank compensation more effective.

\subsection{Error Reconstruction via Whitening SVD}

Firstly, we illustrate how to process quantization error using the whitening technique and establish the mapping between singular values and compression loss. To do this, we first obtain an orthogonal whitened activation $\textbf{S}^{-1}\textbf{X}$ where each channel is independent from each other:

\begin{equation}
    \left(\textbf{S}^{-1} \textbf{X}\right)\left(\textbf{S}^{-1} \textbf{X}\right)^\top=\textbf{S}^{-1} \textbf{X} \textbf{X}^\top\left(\textbf{S}^{-1}\right)^\top=\textbf{I},
\end{equation}

where $\textbf{S}$ can be derived by Cholesky decomposition~\cite{2000Matrix}. Then we perform SVD on $\textbf{E}_q\textbf{S}$, which produces $\textbf{E}_q\textbf{S} = \textbf{U} \mathbf{\Sigma} \textbf{V}^\top$, where $\textbf{U}=[\textbf{u}_1, \textbf{u}_2, ..., \textbf{u}_n]$, $\mathbf{\Sigma}=diag(\sigma_1, \sigma_2, ..., \sigma_n)$ and $\textbf{V}=[\textbf{v}_1, \textbf{v}_2, ..., \textbf{v}_n]^\top$. Suppose we would approximate $\textbf{E}_q$ by rank $r$, the smallest $n-r$ singular values of $\mathbf{\Sigma}$ will be truncated to obtain the compressed $\tilde{\textbf{E}}_q$:

\begin{equation}
\label{equ:cholesky}
    \tilde{\textbf{E}}_q = \textbf{U}_r \mathbf{\Sigma}_r \textbf{V}_r^\top \textbf{S}^{-1}.
\end{equation}

Next, we provide a theoretical justification establishing the connection between singular values and error compensation, showing that the top singular values indeed contribute the most to mitigating the overall error.

For an arbitrary matrix $\textbf{A} \in \mathbb{R}^{m \times m}$, its Frobenius norm can be deduced into the square root of the trace of its gram matrix~\cite{Horn_Johnson_1991}:

\begin{equation}
\label{equ:frobenius}
    \|\textbf{A}\|_F = \left(\sum_{j=1}^n \sum_{i=1}^m\left|a_{i j}\right|^2\right)^{\frac{1}{2}}=\left[\operatorname{Tr}\left(\textbf{A}^\top \textbf{A}\right)\right]^{\frac{1}{2}},
\end{equation}

based on which we can derive the approximation loss $L_i$ of $\tilde{\textbf{E}}_q\textbf{X}$ when truncating the $i^\text{th}$ singular value:

\begin{equation}
\begin{aligned}
L_i
&= \|(\textbf{E}_q-\tilde{\textbf{E}}^i_q)\textbf{X}\|_F
= \|\sigma_i\textbf{u}_i\textbf{v}_i^\top \textbf{S}^{-1}\textbf{X}\|_F \\
&\overset{(i)}{=} \sigma_i \operatorname{Tr}\left(\textbf{u}_i \textbf{v}_i^\top \textbf{S}^{-1} \textbf{X X}^\top(\textbf{S}^{-1}\right)^\top \textbf{v}_i \textbf{u}_i^\top)^{\frac{1}{2}} \\
&\overset{(ii)}{=} \sigma_i \operatorname{Tr}(\textbf{u}_i \textbf{v}_i^\top \textbf{v}_i \textbf{u}_i^\top)^{\frac{1}{2}}
\overset{(iii)}{=}\sigma_i,
\end{aligned}
\end{equation}

where $(i)$ refers to Eq.~\ref{equ:frobenius}, $(ii)$ refers to Eq.~\ref{equ:cholesky} and $(iii)$ equals since $\textbf{v}_i^\top \textbf{v}_i=\textbf{u}_i^\top \textbf{u}_i=I ; \textbf{v}_i^\top \textbf{v}_j=\textbf{u}_i^\top \textbf{u}_j=0, \forall i \neq j ;  \operatorname{Tr}(\textbf{v}_i \textbf{v}_i^\top)= \operatorname{Tr}(\textbf{u}_i \textbf{u}_i^\top)=1$.

Therefore, $\tilde{\textbf{E}}_q$ can be replaced with a pair of LoRA-style matrices $\textbf{L}_A=\textbf{U}_r\mathbf{\Sigma}_r, \textbf{L}_B=\textbf{V}_r^\top \textbf{S}^{-1}$ to compensate for the quantization error of the model. In addition to setting the same rank $r$ for all layers, we use the ratio of the cumulative sum of sorted singular values to the total sum of singular values to set a threshold $\alpha$ whose range is $(0,1)$. 

\begin{equation}
    \max_r\frac{\sum_{i=1}^r\sigma_i}{\sum_{i=1}^n\sigma_i}<\alpha.
\label{equ:rank}
\end{equation}

\subsection{Activation Smoothing via Outlier Analysis}

Based on the empirical observations presented in the analysis section, we find that the outliers play a critical role in error reconstruction. Therefore, we devise a heuristic method to process these outliers. The activation outliers are extracted and migrated to the weights, after which the outlier parts of weights can be split and compensated along with the quantization error reconstruction.

Firstly, inspired by SmoothQuant~\cite{xiao2023smoothquant}, we use a scaling matrix $M$ to migrate the quantization difficulty of the activation to weight:

\begin{equation}
    \textbf{WX} = \textbf{WM}\cdot \textbf{M}^{-1}\textbf{X},
\end{equation}

where $\textbf{M}=diag(\textbf{m}_1, \textbf{m}_2, \dots, \textbf{m}_n)$ is a specially designed matrix. Let $\bar {\textbf{X}}$ and $\bar {\textbf{W}}$ denote the absolute mean activation and weight value of each channel respectively. Suppose $f$ is a pre-defined hyper-parameter, we use $I_f = \{i_1, i_2, i_3, \dots, i_f\}$ to represent the index set of $\bar {\textbf{X}} \odot \bar {\textbf{W}}$ outliers. $\textbf{M}$ can be obtained through the following formula:



\begin{equation}
    \textbf{m}_i = 
    \begin{cases}
        & \bar {\textbf{X}}_i / \bar {\textbf{X}}_{\text{min}}  \qquad i \in I_f \\
        & 1 \quad \qquad \qquad \text{else},
    \end{cases}
\end{equation}

where $\bar {\textbf{X}}_{\text{min}}$ is the minimum value in $\bar {\textbf{X}}_{I_f}$. This makes the activation values easier to quantize and meanwhile allows us to use whitening SVD to uniformly treat and compensate for integral quantization error in the weight matrix:

\begin{equation}
\begin{aligned}
\textbf{WM}\cdot \textbf{M}^{-1}\textbf{X}
&= (\textbf{W}_s + \textbf{W}_o)\textbf{M}^{-1}\textbf{X} \\
&= (Q(\textbf{W}_s) + (\textbf{E}_q + \textbf{W}_o))\textbf{M}^{-1}\textbf{X}.
\end{aligned}
\end{equation}

The outlier part $\textbf{W}_o$ will be extracted from the adjusted weight $\textbf{WM}$ based on the outlier index set $I_f$, and it will not be quantized. Together with the quantization errors from the rest of the weights, they are compensated by low-rank approximation matrices to minimize the loss incurred during the quantization process. Thus, we morph the objective of the whitening SVD into $(\textbf{E}_q+\textbf{W}_o)$, and similar narrow matrices are used to approximate this error:

\begin{equation}
    \textbf{L}_A\textbf{L}_B \sim (\textbf{E}_q+\textbf{W}_o)\textbf{S}.
\end{equation}

In this way, the inference pass is replaced by $\textbf{W}_q\textbf{X}+\textbf{L}_A\textbf{L}_B\textbf{X}$ for error reconstruction. Note that this technique is orthogonal to the weight-only quantization, which means ASER can be conducted based on the existing weight-only quantization methods, not limited to RTN. To sum up, the pseudo-code of our algorithm is as follows:

\begin{algorithm}[h]
\caption{Activation Smoothing and Error Reconstruction for Post-Training Quantization}
\label{alg:aser}
\textbf{Input}: Original weight $\textbf{W}$, Calibration data $\mathcal{D}$ \\
\textbf{Parameter}: Rank threshold $\alpha$, Outlier threshold $f$ \\
\textbf{Output}: Quantized Weight $\textbf{W}_q$, Reconstruction matrices $\textbf{L}_A, \textbf{L}_B$ over $L$ layers
\begin{algorithmic}[1] 
\FOR {Data $d$ in $\mathcal{D}$}
\STATE Inference data $d$ and store activations $\textbf{X}$
\ENDFOR

\FOR {Layer $\textbf{W}^l$ in $\textbf{W}$}
\STATE Intialize $\textbf{m} = [1, 1, ..., 1]^n$ and  $\textbf{M} = diag(\textbf{m})$
\IF{with Activation Smoothing~(A.S.)}
\STATE Compute $\textbf{m}_{i}= \bar {\textbf{X}}_i^l / \bar {\textbf{X}}_{\text{min}}^l, \text{for}\ i \in I_f$
\STATE Split $\textbf{W}^l\textbf{M} = \textbf{W}_s^l + \textbf{W}_o^l$
\STATE Compute $\textbf{E}_q^l = \textbf{W}^l - Q(\textbf{W}_s^l) + \textbf{W}_o^l$
\ELSE
\STATE Compute $\textbf{E}_q^l = \textbf{W}^l - Q(\textbf{W}^l)$
\ENDIF
\STATE Compute $\textbf{S}$ by Cholesky decomposition on $(\textbf{M}^{-1}\textbf{X})(\textbf{M}^{-1}\textbf{X})^\top$: 
$(\textbf{S}^{-1}\textbf{M}^{-1}\textbf{X})(\textbf{S}^{-1}\textbf{M}^{-1}\textbf{X})^\top = \textbf{I}$
\STATE SVD decomposition: $\textbf{U}\mathbf{\Sigma} \textbf{V}^\top = \textbf{E}_q^l\textbf{S}$
\STATE Compute $r$ according to Eq.~\ref{equ:rank}
\STATE Compute $\textbf{L}_A^l=\textbf{U}_r\mathbf{\Sigma}_r, \textbf{L}_B^l=\textbf{V}_r^\top \textbf{\textbf{S}}^{-1}$
\ENDFOR
\STATE \textbf{return} $\{\textbf{W}^l_q, \textbf{L}^l_A, \textbf{L}^l_B\}_{l=1}^L$
\end{algorithmic}
\end{algorithm}

\section{Experiments}

This section presents experimental results by different PTQ methods on various model architectures. We further conduct visualization analysis to validate the effectiveness of the method, followed by an overhead analysis.

\subsection{Experimental Setup}

\noindent\textbf{Models} We consider quantization of typical open source large language models involving LLaMA3-8B~\cite{touvron2023llama}, Qwen1.5-7B, Qwen-72B~\cite{bai2023qwen}. We obtain the pre-trained model from the official repository\footnote{https://github.com/meta-llama/llama}\footnote{https://github.com/QwenLM/Qwen}.

\noindent\textbf{Baselines} We consider several popular PTQ algorithms for comparison. The baseline approaches include
act-and-weight quantization approaches LLM.int()~\cite{dettmers2022llmint88bitmatrixmultiplication},  SmoothQuant~\cite{xiao2023smoothquant} and SmoothQuant+~\cite{pan2023smoothquantaccurateefficient4bit}. Furthermore, we compared the quantization algorithm LoRC~\cite{yao2024exploring}, L$^2$QER~\cite{zhang2024lqer} using additional parameter. In our results, we use w/ and w/o A.S. to indicate whether the activation smoothing has been employed.

\noindent\textbf{Evaluation Metrics} Firstly, we evaluate quantized model by perplexity~(PPL) on language modeling datasets Wikitext2~\cite{merity2016pointer}, PTB~\cite{marcus1994penn} and a portion of the C4 dataset~\cite{raffel2020exploring}. Then we conduct experiments on zero-shot evaluation tasks including ARC-e, ARC-c~\cite{clark2018think}, MMLU~\cite{hendrycks2020measuring}, HellaSwag~\cite{zellers2019hellaswag}, PIQA~\cite{bisk2020piqa}, GSM8K~\cite{cobbe2021gsm8k} and Human-Eval~\cite{chen2021codex}. We experiment on the open source evaluation framework OpenCompass~\cite{2023opencompass}.

\noindent\textbf{Experimental Details} To ensure a fair comparison in our assessment, we standardize the usage of the calibration dataset for all quantization methods. All quantization is conducted by per-channel for weights and per-token for activations. We consistently employ a sample size of $128$ and maintain a uniform token sequence length of $2048$ for calibration throughout the evaluation process. All experiments are conducted on $4\times$ NVIDIA A100 GPUs.

\begin{table*}[t]
\centering
\caption{Evaluation results of post-training quantization on LLaMA3-8B. Best results are marked in bold.}
\begin{tabular}{lllrrrrrrrrr}
\hline
\multicolumn{1}{l|}{\multirow{2}{*}{\textbf{Method}}} & \multirow{2}{*}{\textbf{\#W}} & \multicolumn{1}{l|}{\multirow{2}{*}{\textbf{\#A}}} & \multicolumn{3}{c|}{\textbf{Perplexity$(\downarrow)$}} & \multicolumn{6}{c}{\textbf{Accuracy$(\uparrow)$}} \\ \cline{4-12} 
\multicolumn{1}{l|}{} &  & \multicolumn{1}{l|}{} & \textbf{WikiText2} & \textbf{C4} & \multicolumn{1}{r|}{\textbf{PTB}} & \textbf{ARC-e} & \textbf{ARC-c} & \textbf{MMLU} & \textbf{Hella} & \textbf{PIQA} & \textbf{Avg.} \\ \hline
LLaMA3-8B & 16 & 16 & 6.14 & 10.75 & 9.91 & 89.77 & 79.66 & 37.85 & 48.47 & 68.55 & 64.86 \\ \hline \hline
LLM.int4() & 4 & 8 & 10.57 & 18.43 & 15.52 & 77.07 & 59.32 & 13.48 & 28.83 & 54.68 & 46.68 \\
SmoothQuant & 4 & 8 & 11.23 & 18.87 & 16.78 & 60.32 & 44.75 & 14.99 & 26.93 & 37.65 & 36.93 \\
SmoothQuant+ & 4 & 8 & 10.22 & 17.39 & 15.34 & 72.31 & 55.93 & 14.25 & 26.44 & 55.17 & 44.82 \\
LoRC & 4 & 8 & 8.62 & 15.09 & 12.99 & 83.07 & 65.08 & 10.38 & 29.79 & 60.99 & 49.86 \\
L$^2$QER & 4 & 8 & 7.72 & 13.51 & 11.84 & 86.07 & 69.15 & 26.83 & 30.42 & \textbf{61.97} & 54.89 \\ \hline
ASER (w/o A.S.) & 4 & 8 & 7.61 & 13.24 & 11.65 & \textbf{86.24} & 72.20 & 27.90 & \textbf{31.08} & 59.47 & 55.38 \\
ASER (w/ A.S.) & 4 & 8 & \textbf{7.43} & \textbf{12.94} & \textbf{11.36} & \textbf{86.24} & \textbf{72.54} & \textbf{30.73} & 29.62 & 60.50 & \textbf{55.93} \\  \hline \hline
LLM.int4() & 4 & 6 & 14.21 & 24.52 & 22.89 & 51.85 & 40.68 & 8.30 & 24.54 & 49.18 & 34.91 \\
SmoothQuant & 4 & 6 & 15.74 & 25.29 & 24.42 &  39.51 & 30.17 & 7.58 & 23.17 & 41.73 & 28.43 \\
SmoothQuant+ & 4 & 6 & 29.22 & 39.10 & 42.46 &  30.16 & 32.54 & 15.51 & 19.66 & 41.64 & 27.90 \\
LoRC & 4 & 6 & 11.03 & 19.28 & 16.99 & 61.20 & 42.04 & 3.43 & 25.94 & 54.35 & 37.39 \\
L$^2$QER & 4 & 6 & 9.44 & 16.43 & 14.12 & 70.37 & 56.27 & 10.98 & 26.48 & 54.30 & 43.68 \\ \hline
ASER (w/o A.S.) & 4 & 6 & 9.02 & 15.55 & 13.46 & 73.37 & 52.20 & 7.78 & 26.99 & 55.55 & 43.18 \\
ASER (w/ A.S.) & 4 & 6 & \textbf{8.41} & \textbf{14.63} & \textbf{12.58} & \textbf{78.66} & \textbf{61.69} & \textbf{17.32} & \textbf{27.07} & \textbf{58.00} & \textbf{48.55} \\ \hline
\end{tabular}
\label{tab:llama}
\end{table*}

\begin{table*}[h]
\centering
\caption{Evaluation results of post-training quantization on Qwen1.5-7B. Best results are marked in bold.}
\begin{tabular}{lllrrrrrrrrr}
\bottomrule
\multicolumn{1}{l|}{\multirow{2}{*}{\textbf{Method}}} & \multirow{2}{*}{\textbf{\#W}} & \multicolumn{1}{l|}{\multirow{2}{*}{\textbf{\#A}}} & \multicolumn{3}{c|}{\textbf{Perplexity$(\downarrow)$}} & \multicolumn{6}{c}{\textbf{Accuracy$(\uparrow)$}} \\ \cline{4-12} 
\multicolumn{1}{l|}{} &  & \multicolumn{1}{l|}{} & \textbf{WikiText2} & \textbf{C4} & \multicolumn{1}{l|}{\textbf{PTB}} & \textbf{ARC-e} & \textbf{ARC-c} & \textbf{MMLU} & \textbf{Hella} & \textbf{PIQA} & \textbf{Avg.} \\ \hline
Qwen1.5-7B & 16 & 16 & 7.95 & 13.57 & 11.94 & 87.48 & 76.61 & 49.78 & 61.49 & 66.76 & 68.42 \\ \hline \hline
LLM.int4() & 4 & 8 & 23.73 & 38.14 & 25.28 & 17.46 & 14.58 & 13.27 & 0.18 & 12.08 & 11.51 \\
SmoothQuant & 4 & 8 & 15.73 & 28.18 & 21.24 & 33.51 & 26.44 & 24.00 & 1.88 & 31.94 & 21.46 \\
SmoothQuant+ & 4 & 8 & 46.27 & 51.28 & 40.81 & 28.04 & 22.03 & 9.16 & 1.17 & 29.60 & 18.00 \\
LoRC & 4 & 8 & 16.78 & 27.13 & 19.93 & 76.19 & 66.44 & 21.56 & 43.77 & 45.43 & 50.68 \\
L$^2$QER & 4 & 8 & 9.20 & 15.63 & 13.81 & \textbf{85.36} & 71.53 & 46.39 & 29.59 & 64.64 & 59.50 \\ \hline
ASER (w/o A.S.) & 4 & 8 & 9.19 & 15.59 & 13.69 & 81.83 & 68.81 & 43.89 & 45.44 & 61.10 & 60.21 \\
ASER (w/ A.S.) & 4 & 8 & \textbf{8.72} & \textbf{14.84} & \textbf{13.10} & 84.66 & \textbf{72.54} & \textbf{49.55} & \textbf{52.18} & \textbf{67.57} & \textbf{65.30} \\
\hline \hline
LLM.int4() & 4 & 6 & 34.32 & 56.00 & 39.67 & 50.09  & 39.66 & 20.89 & 16.11 & 35.69 & 32.49 \\
SmoothQuant & 4 & 6 & 26.81 & 42.67 & 35.62 & 41.98 & 35.59 & 16.08 & 7.31 & 32.59 & 26.71 \\
SmoothQuant+ & 4 & 6 & 131.45 & 176.46 & 130.24 & 35.98 & 30.51 & 17.36 & 18.77 & 33.35 & 27.19 \\
LoRC & 4 & 6 & 23.03 & 39.63 & 29.68 & 45.68 & 37.63 & 8.90 & 26.29 & 30.85 & 29.87 \\
L$^2$QER & 4 & 6 & 16.51 & 27.01 & 22.46 & 63.14 & 47.80 & 21.85 & 30.24 & 51.52 & 42.91 \\ \hline
ASER (w/o A.S.) & 4 & 6 & 15.86 & 26.38 & 21.25 & 67.37 & 54.92 & \textbf{40.34} & 30.56 & \textbf{54.57} & 49.52 \\
ASER (w/ A.S.) & 4 & 6 & \textbf{11.03} & \textbf{17.57} & \textbf{16.13} & \textbf{75.31} & \textbf{62.37} & 32.07 & \textbf{34.49} & 43.63 & \textbf{49.57} \\ \hline
\end{tabular}
\label{tab:qwen}
\end{table*}

\subsection{Main Results}

We demonstrate the main results of LLaMA3-8B and Qwen1.5-7B evaluation in Tables~\ref{tab:llama} and \ref{tab:qwen} respectively. The methods are divided into two parts, W4A8 and W4A6 quantization setups. We both test the perplexity and accuracy, and the best results are marked in bold. To ensure a fair comparison, we set the rank of compensation parameters in each layer of ASER, LoRC, and L$^2$QER to $64$. The outlier threshold $f$ is set by $32$, which is equal in experimental setup.

As shown in the table~\ref{tab:llama}, ASER achieves the best perplexity on language modeling tasks, while simultaneously performs competitive accuracy on language understanding. In both settings, ASER achieves perplexity improvement compared to the fp16 baseline ranging from $1.20$ to $3.88$. For the accuracy evaluation, particularly in the W4A6 per-channel setting, existing algorithms without error compensation, e.g., SmoothQuant and its variant, show significant performance degradation, while ASER outperforms its counterparts L$^2$QER by $4.87\%$, which attributes to the significant role of activation smoothing.

In Table~\ref{tab:qwen}, for the W4A8 quantization, ASER shows competitive performance compared to the state-of-the-art method, with $3.12\%$ close to the benchmark accuracy of the fp16 baseline averagely. Thanks to the activation smoothing, ASER achieves $5.09\%$ higher accuracy compared to the one with only error reconstruction. For W4A6 quantization setup on language modeling datasets, as we can see, many existing methods fail to keep the performance without group-wise scaling. However, the quantized model using ASER achieves the perplexity that is $3.08\sim 4.19$ close to the fp16 baseline. Similarly, it also delivers the best accuracy on the commonsense evaluation dataset. Particularly, compared to L$^2$QER, ASER performs better accuracy by $6.66\%$, which also far exceeds other act-and-weight quantization baselines.

\begin{table}[]
\centering
\caption{Evaluation results on Qwen-72B. All models are quantized by W4A8 per-channel setup. The baseline model is paramterized by fp16. HEval represents the Human-Eval benchmark. Best results are marked in bold.}
\begin{tabular}{lrrrr}
\hline
\multicolumn{1}{l|}{\multirow{2}{*}{\textbf{Method}}} & \multicolumn{4}{c}{\textbf{Accuracy$(\uparrow)$}} \\ \cline{2-5} 
\multicolumn{1}{l|}{} & \multicolumn{1}{r}{\textbf{ARC-e}} & \multicolumn{1}{r}{\textbf{ARC-c}} & \multicolumn{1}{r}{\textbf{GSM8K}} & \multicolumn{1}{r}{\textbf{HEval}} \\ \hline
Qwen-72B & 94.36 & 88.47 & 70.13 & 50.61 \\ \hline \hline
LLM.int4() & 92.77 & 84.75 & 56.33 & 39.02 \\
SmoothQuant & 72.13 & 65.08 & 56.71 & 38.41 \\
SmoothQuant+ & 86.07 & 77.63 & 60.58 & 37.20 \\
LoRC & 93.84 & \textbf{88.14} & 64.06 & 45.12 \\
L$^2$QER & 92.95 & 85.46 & 66.19 & 42.07 \\ \hline
ASER (w/o AS) & \textbf{94.00} & \textbf{88.14} & 64.14 & 44.51 \\
ASER (w/ AS) & 92.59 & 85.76 & \textbf{67.32} & \textbf{46.34} \\ \hline

\end{tabular}
\label{tab:qwen72}
\end{table}


\begin{figure*}[t]
\centering
\includesvg[width=2\columnwidth]{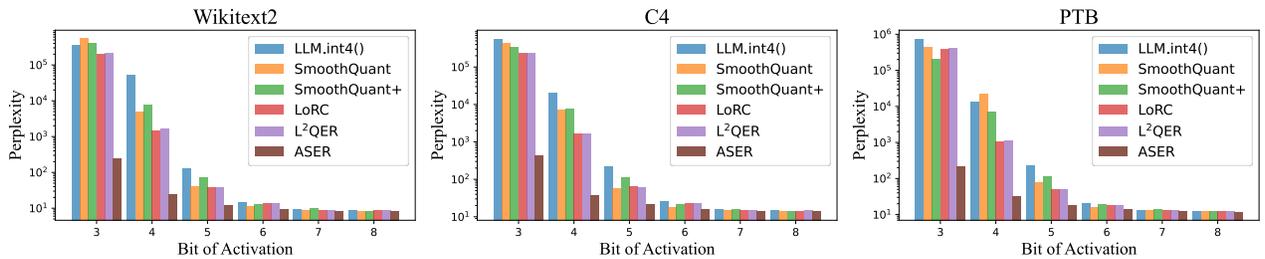}
\caption{Perplexity of quantized Qwen1.5-7B with int8 weight and different quantization bit of activation, i.e., W8Ax.}
\label{fig:ppl_bit}
\end{figure*}

\subsection{Ablation Study and Analysis}

\noindent\textbf{Results of Larger Model} 
Next, we validate the effectiveness of our method on the models with larger-scale. The model involved is Qwen-72B in Table~\ref{tab:qwen72}. We mainly evaluate the setup of W4A8 per-channel quantization on commonsense benchmarks. As we can see, relying on our low-rank quantization error reconstruction algorithm, with further activation smoothing, ASER achieves an accuracy of $0.33\%\sim 4.27\%$, which is close to the fp16 baseline. This demonstrates the scalability of our approach and its adaptability to larger models. Experimental results of more quantization setups and models and can be found in Appendix.

\noindent\textbf{Effect of Activation Smoothing} As previously mentioned, ASER employs several techniques, including activation smoothing, migration of outliers, and low-rank estimation to compensate for integral quantization errors. These techniques make activation quantization more favorable. The results of the main experiments demonstrate the improvement brought by activation smoothing particularly in activation quantization setup. So we further conduct some experiments to study this effect. As Figure~\ref{fig:ppl_bit} shows, the existing quantization algorithms reach intolerable perplexity with low bit activation. LLM.int4(), SmoothQuant and its variant, cannot withstand the performance degradation caused by low activation bit-width. Though LoRC and L$^2$QER do compensation for the quantized weight, they are not able to process activation outliers. However, ASER keeps remarkably better performance even in 4-bit activation quantization.

\begin{figure}[]
\centering
\includesvg[width=\columnwidth]{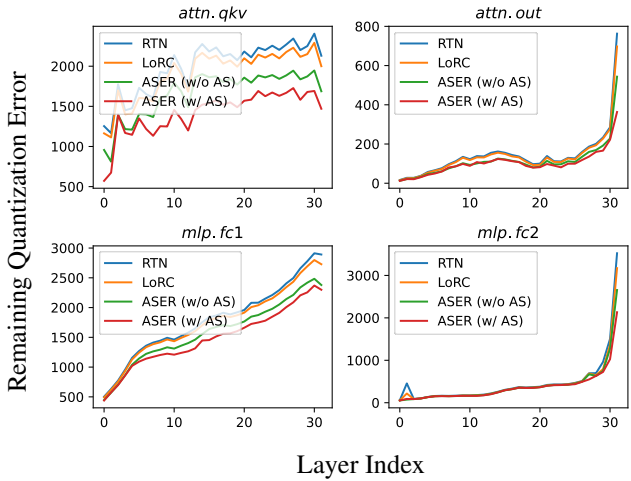}
\caption{The remaining quantization error across layers of LLaMA3-8B using different methods of W4A6 quantization. The error is $\|\textbf{W}
\textbf{X}-\textbf{W}_q\textbf{X}_q\|_F$ for the RTN, while the other methods represent the error norm after reconstruction.}
\label{fig:error}
\end{figure}

\noindent\textbf{Effect of Error Compensation} We visualize the compensation performance of different methods in Fig.~\ref{fig:error}, where the remaining quantization errors of four consecutive layers are plotted over transformer layers in LLaMA3-8B. The RTN represents the baseline quantization error without any compensation. The figure demonstrates that ASER can effectively reconstruct the quantization error, compared to LoRC. Furthermore, the application of activation smoothing enhances the error reconstruction capability, due to the extraction and processing of the outliers.

\noindent\textbf{Rank Selection} In the ASER framework, the choice of rank $r$, or its threshold $\alpha$, influences the amount of quantization error compensation and the additional number of parameters introduced. Therefore, we conduct an experimental analysis on how the selection of rank affects downstream tasks. For Qwen1.5-7B, the performance of downstream tasks positively correlates with the selection of $\alpha$, whereas this might not be the case for other models. As shown in Table~\ref{tab:rank}, we find that using a larger rank is not necessarily better for all tasks in Qwen-7B. When we set smaller rank, we might achieve a better balance between performance of W4A8-quantized model and the extra parameter count.

\begin{table}[h]
\centering
\caption{Zero-shot accuracy evaluation results and overhead of Qwen-7B using ASER with different rank configuration.}
\begin{tabular}{rrrrrr}
\hline
$\alpha$ & $\bar r$ & ARC-e & Hella & PIQA & + FLOPs \\ \hline
0.1 & 59.93 & 83.25 & \textbf{68.39} & 72.52 & 2.93\% \\
0.075 & 49.60 & \textbf{84.48} & 67.63 & 72.74 & 2.42\% \\
0.05 & 30.58 & 82.36 & 67.33 & 72.80 & 1.49\% \\
0.03 & 14.00 & 80.78 & 67.41 & \textbf{72.96} & 0.68\% \\
0.015 & 5.41 & 82.89 & 67.01 & 71.98 & 0.26\% \\ \hline
\end{tabular}
\label{tab:rank}
\end{table}

\subsection{Overhead Analysis} The proposed method can introduce additional computations and memory overheads due to the involvement of extra LoRA-style matrices $\textbf{L}_A, \textbf{L}_B$. Let $s$ denote the sequence length, $d$ the hidden dimension, $r(\ll d)$ the rank. The complexity analysis of a single layer is followed, where the computation represents floating-point operations~(FLOPs), and the memory is measured by the number of parameters:

\begin{table}[h]
\centering
\begin{tabular}{lll}
\hline
Complexity & Origin & ASER \\ \hline
Computation & $sd^2$ & $sd^2+2srd$ \\
Memory & $d^2$ & $d^2+2rd$ \\ \hline
\end{tabular}
\end{table}

Taking Qwen-7B~($s=2048, d=4096$) as an example, we demonstrate the tradeoff of accuracy and the computational overhead among the listed configurations of Table~\ref{tab:rank}. The choice can be deliberated based on the model generality or specific tasks. The additional parameter introduces minimum theoretical computational overhead of only 0.26\%, which is minor to the main network inference. We leave it for future work to adaptively choose the best $\alpha$.

\section{Conclusion}

In this work, from the perspective of model compression, we formulate and analyze the quantization error. Our study reveals that the error exhibits low-rank property, whose major components originate from the combined effect of outliers in both activations and weights. In response to these findings, we propose ASER, which utilizes whitening SVD to compensate for the integral error~(Error Reconstruction), followed by analyzing and extracting activation outliers~(Activation Smoothing). Experimental results demonstrate that ASER outperforms existing quantization algorithms on typical large language models in various quantization setups, with negligible theoretical overhead.

\newpage

\bibliography{aaai25}

\end{document}


\appendix

\maketitle

\section{Numerical Effect of Activation Smoothing}

The idea of activation smoothing is to extract the outliers composed by both weights and activations, and transfer the difficulty of activation quantization, then the outliers part will be construct within the quantization error reconstruction. In Fig.~\ref{fig:smooth}, we demonstrate that, in the numerical perspective, the activation range is significantly smoothed, while the outliers transfered into the weight are extracted by $W_o$.

\begin{figure}[h]
\centering
\includesvg[width=\columnwidth]{figures/smooth.svg}
\caption{The effect of activation smoothing. The figures on the left and right respectively represent $A$ and $W$ before and after smoothing, where $WM=W_s+W_o$. The data comes from the first layer in Qwen1.5-7B.}
\label{fig:smooth}
\end{figure}

\section{Effect of Rank Selection Strategy}

As mentioned before, we observe the different effective dimentionality of quantization error across the layers in LLM. When it comes to the construction of LoRA-style matrix to compensate for the error, we design the strategy to dynamically select the rank configuration. As Fig.~\ref{fig:rank} shows, when using $\alpha$ ranging from $0.015$ to $0.1$, we employ different ranks across layers, whose trends are relatively consistent within the same layer across the Transformer blocks. The distribution also conforms to the effective rank in our analysis of quantization error.

\begin{figure}[h]
\centering
\includesvg[width=\columnwidth]{figures/rank.svg}
\caption{The effect of rank selection. The rank is set by different threshold $\alpha$ in different layers from LLaMA3-8B.}
\label{fig:rank}
\end{figure}

\section{More Evaluation Results}

We extensively evaluate ASER on models of various scales and types, and on different datasets, to demonstrate its generalization ability. Specifically, we supplements the weight-only quantization on LLaMA3-8B~(Table~\ref{tab:llama8b}). Compared to the state-of-the-art approaches GPTQ and AWQ, ASER achieves competitive results both in causal language modeling perplexity and commonsense tasks accuracy. Similar results of weight-only quantization setup are shown in LLaMA2-13B~(Table~\ref{tab:llama13b}).

Note that, with activation smoothing, the advantage of ASER lies in its quantization of activations. Therefore, we emphasize testing on act-and-weight quantization. Besides LLaMA2-13B~(Table~\ref{tab:llama13b}), larger models are involved in the evaluation. As for Qwen-14B~(Table~\ref{tab:qwen14b}), Qwen1.5-32B~(Table~\ref{tab:qwen32b}), ASER shows close average accuracy related to the fp16 baseline model in the W4A8 per-channel quantization setup.

\begin{table*}[ht]
\centering
\caption{Evaluation results of post-training quantization on LLaMA3-8B. Best results are marked in bold.}
\begin{tabular}{lllrrrrrrrrr}
\hline
\multicolumn{1}{l|}{\multirow{2}{*}{\textbf{Method}}} & \multirow{2}{*}{\textbf{\#W}} & \multicolumn{1}{l|}{\multirow{2}{*}{\textbf{\#A}}} & \multicolumn{3}{c|}{\textbf{Perplexity$(\downarrow)$}} & \multicolumn{6}{c}{\textbf{Accuracy$(\uparrow)$}} \\ \cline{4-12} 
\multicolumn{1}{l|}{} &  & \multicolumn{1}{l|}{} & \textbf{WikiText2} & \textbf{C4} & \multicolumn{1}{r|}{\textbf{PTB}} & \textbf{ARC-e} & \textbf{ARC-c} & \textbf{MMLU} & \textbf{Hella} & \textbf{PIQA} & \textbf{Avg.} \\ \hline
LLaMA3-8B & 16 & 16 & 6.14 & 10.75 & 9.91 & 89.77 & 79.66 & 37.85 & 48.47 & 68.55 & 64.86 \\ \hline \hline
RTN & 4 & 16 & 10.21 & 17.86 & 15.09 & 79.01 & 60.68 & 12.89 & 30.85 & 55.82 & 47.85 \\
GPTQ & 4 & 16 & 7.91 & 13.81 & 11.86 & 79.89 & 66.78 & 26.43 & 36.07 & 61.32 & 54.10 \\
AWQ & 4 & 16 & 7.86 & 13.70 & 11.74 & 85.36 & 72.88 & 20.47 & \textbf{36.26} & \textbf{60.99} & 55.19 \\ \hline
ASER (w/o AS) & 4 & 16 & \textbf{7.34} & \textbf{12.74} & \textbf{11.20} & 86.77 & \textbf{74.24} & 28.54 & 31.51 & 58.87 & 55.99 \\
ASER (w/ AS) & 4 & 16 & 7.44 & 12.94 & 11.40 & \textbf{86.95} & 72.88 & \textbf{31.20} & 29.39 & 60.55 & \textbf{56.19} \\ \hline
\end{tabular}
\label{tab:llama8b}
\end{table*}


\begin{table*}[ht]
\centering
\caption{Evaluation results of post-training quantization on LLaMA2-13B. Best results are marked in bold.}
\begin{tabular}{lllrrrrrrrrr}
\hline
\multicolumn{1}{l|}{\multirow{2}{*}{\textbf{Method}}} & \multirow{2}{*}{\textbf{\#W}} & \multicolumn{1}{l|}{\multirow{2}{*}{\textbf{\#A}}} & \multicolumn{3}{c|}{\textbf{Perplexity$(\downarrow)$}} & \multicolumn{6}{c}{\textbf{Accuracy$(\uparrow)$}} \\ \cline{4-12} 
\multicolumn{1}{l|}{} &  & \multicolumn{1}{l|}{} & \textbf{WikiText2} & \textbf{C4} & \multicolumn{1}{l|}{\textbf{PTB}} & \textbf{ARC-e} & \textbf{ARC-c} & \textbf{MMLU} & \textbf{Hella} & \textbf{PIQA} & \textbf{Avg.} \\ \hline
LLaMA2-13B & 16 & 16 & 4.88 & 7.55 & 34.42 & 68.08 & 48.47 & 45.42 & 31.92 & 55.88 & 49.95 \\ \hline \hline
RTN & 4 & 16 & 5.39 & 8.21 & 46.13 & 64.20 & 46.78 & 37.21 & 29.36 & \textbf{57.62} & 47.03 \\
GPTQ & 4 & 16 & 5.22 & 8.05 & 74.85 & 63.32 & \textbf{52.88} & 42.40 & 31.02 & 53.81 & 48.69 \\
AWQ & 4 & 16 & 5.33 & 8.14 & 50.17 & 59.96 & 42.37 & 40.27 & 28.44 & 55.71 & 45.35 \\ \hline
ASER (w/o A.S.) & 4 & 16 & 5.15 & 7.85 & 39.09 & \textbf{68.43} & 48.81 & \textbf{47.30} & \textbf{31.96} & 54.90 & \textbf{50.28} \\
ASER (w/ A.S.) & 4 & 16 & \textbf{5.12} & \textbf{7.85} & \textbf{38.65} & 67.37 & 50.51 & 44.82 & 31.02 & 55.88 & 49.92 \\ \hline
LLM.int4() & 4 & 8 & 5.46 & 8.30 & 48.77 & 63.49 & 44.75 & 36.39 & 28.75 & 57.45 & 46.17 \\
SmoothQuant & 4 & 8 & 6.10 & 9.26 & 52.60 & 39.66 & 22.75 & 30.03 & 25.26 & 50.44 & 33.63 \\
LoRC & 4 & 8 & 5.28 & 8.03 & \textbf{38.45} & 65.08 & \textbf{50.17} & 44.83 & 30.11 & 57.07 & 49.45 \\
L$^2$QER & 4 & 8 & 5.21 & 7.95 & 39.57 & 67.20 & 48.81 & 45.19 & 30.09 & 56.64 & 49.59 \\ \hline
ASER (w/o A.S.) & 4 & 8 & 5.21 & 7.92 & 40.45 & \textbf{68.08} & 48.14 & \textbf{47.81} & \textbf{31.22} & \textbf{58.16} & \textbf{50.68} \\
ASER (w/ A.S.) & 4 & 8 & \textbf{5.16} & \textbf{7.89} & 39.44 & 67.02 & 49.83 & 45.08 & 29.83 & 57.56 & 49.86 \\
\hline
\end{tabular}
\label{tab:llama13b}
\end{table*}

\newpage

\begin{table*}[ht]
\centering
\caption{Evaluation results of post-training quantization on Qwen-14B. Best results are marked in bold.}
\begin{tabular}{llllllll}
\hline
\multicolumn{1}{l|}{\multirow{2}{*}{\textbf{Method}}} & \multirow{2}{*}{\textbf{\#W}} & \multicolumn{1}{l|}{\multirow{2}{*}{\textbf{\#A}}} & \multicolumn{5}{c}{\textbf{Accuracy$(\uparrow)$}} \\ \cline{4-8} 
\multicolumn{1}{l|}{} &  & \multicolumn{1}{l|}{} & \multicolumn{1}{r}{\textbf{ARC-e}} & \multicolumn{1}{r}{\textbf{ARC-c}} & \multicolumn{1}{r}{\textbf{Hella}} & \multicolumn{1}{r}{\textbf{PIQA}} & \multicolumn{1}{r}{\textbf{Avg.}} \\ \hline
Qwen-14B & 16 & 16 & 92.24 & 83.05 & 80.42 & 73.88 & 82.40 \\ \hline \hline
LLM.int4() & 4 & 8 & 86.77 & 79.32 & 41.31 & 52.99 & 65.10 \\
SmoothQuant & 4 & 8 & 55.73 & 45.08 & 31.33 & 49.18 & 45.33 \\
LoRC & 4 & 8 & \textbf{92.06} & \textbf{84.41} & 65.31 & 67.08 & 77.22 \\
L$^2$QER & 4 & 8 & 91.01 & 83.05 & 71.68 & 69.42 & 78.79 \\ \hline
ASER (w/o A.S.) & 4 & 8 & 91.36 & 82.03 & 71.94 & 70.08 & 78.85 \\
ASER (w/ A.S.) & 4 & 8 & 91.71 & 82.03 & \textbf{74.15} & \textbf{70.84} & \textbf{79.68} \\ \hline
\end{tabular}
\label{tab:qwen14b}
\end{table*}

\begin{table*}[ht]
\centering
\caption{Evaluation results of post-training quantization on Qwen1.5-32B. Best results are marked in bold.}
\begin{tabular}{lllrrrrr}
\hline
\multicolumn{1}{l|}{\multirow{2}{*}{\textbf{Method}}} & \multirow{2}{*}{\textbf{\#W}} & \multicolumn{1}{l|}{\multirow{2}{*}{\textbf{\#A}}} & \multicolumn{5}{c}{\textbf{Accuracy$(\uparrow)$}} \\ \cline{4-8} 
\multicolumn{1}{l|}{} &  & \multicolumn{1}{l|}{} & \textbf{ARC-e} & \textbf{ARC-c} & \textbf{Hella} & \textbf{PIQA} & \textbf{Avg.} \\ \hline
Qwen1.5-32B & 16 & 16 & 84.30 & 71.53 & 16.26 & 56.96 & 57.26 \\ \hline \hline
LLM.int4() & 4 & 8 & 68.61 & 51.19 & 6.88 & 44.78 & 42.86 \\
SmoothQuant & 4 & 8 & 69.14 & 53.90 & 6.10 & 47.17 & 44.08 \\
LoRC & 4 & 8 & 81.83 & 69.49 & \textbf{16.21} & 51.69 & 54.81 \\
L$^2$QER & 4 & 8 & 82.89 & 70.17 & 15.81 & 54.35 & 55.81 \\ \hline
ASER (w/o A.S.) & 4 & 8 & \textbf{85.19} & 69.15 & 14.66 & \textbf{54.84} & 55.96 \\
ASER (w/ A.S.) & 4 & 8 & 85.01 & \textbf{73.56} & 15.08 & 54.52 & \textbf{57.04} \\ \hline
\end{tabular}
\label{tab:qwen32b}
\end{table*}
